\documentclass[conference,a4paper]{IEEEtran}
\IEEEoverridecommandlockouts
\usepackage{cite}
\usepackage{algorithmic}
\usepackage{textcomp}
\usepackage{xcolor}
\usepackage{graphicx}
\usepackage{multirow}
\usepackage{url}
\usepackage{comment}
\usepackage{todonotes}
\usepackage{amsmath, amsthm, amssymb, latexsym, amsfonts}
\usepackage{multicol}
\usepackage{subfig}
\usepackage{booktabs}
\usepackage{csquotes}
\usepackage{float}
\usepackage{hyperref}

\def\BibTeX{{\rm B\kern-.05em{\sc i\kern-.025em b}\kern-.08em
    T\kern-.1667em\lower.7ex\hbox{E}\kern-.125emX}}

\begin{document}

\title{Visual Question Answering in the Medical Domain}

\author{%
  Louisa Canepa\IEEEauthorrefmark{1},
  Sonit Singh\IEEEauthorrefmark{1},
  Arcot Sowmya\IEEEauthorrefmark{1}
  \\
  \IEEEauthorblockA{%
    \IEEEauthorrefmark{1} School of Computer Science and Engineering, University of New South Wales, Sydney, Australia
  }
  \{l.canepa, sonit.singh, a.sowmya\}@unsw.edu.au
}

\maketitle

\begin{abstract}
Medical visual question answering (Med-VQA) is a machine learning task that aims to create a system that can answer natural language questions based on given medical images. Although there has been rapid progress on the general VQA task, less progress has been made on Med-VQA due to the lack of large-scale annotated datasets. In this paper, we present domain-specific pre-training strategies, including a novel contrastive learning pretraining method, to mitigate the problem of small datasets for the Med-VQA task. We find that the model benefits from components that use fewer parameters. We also evaluate and discuss the model's visual reasoning using evidence verification techniques. Our proposed model obtained an accuracy of 60\% on the VQA-Med 2019 test set, giving comparable results to other state-of-the-art Med-VQA models. 
\end{abstract}



\section{Introduction}~\label{Introduction}

With recent advancements in the field of Computer Vision (CV) and Natural Language Processing (NLP), researchers have started looking at \emph{cross-modal} problems that require deeper understanding of both images and text. Of the various tasks at the intersection of CV and NLP, Visual Question Answering (VQA)~\cite{AAL+15} involves taking as input a natural language question and an image, and producing the correct natural language answer. The goal is to design Artificial Intelligence (AI) systems that can form a holistic understanding of images, and are able to effectively express that understanding in natural language. Inspired by the general domain VQA, Medical Visual Question Answering (Med-VQA) takes as input a natural language question and a medical image, and produces a plausible correct natural language answer as the output. The Med-VQA task has gained popularity more recently after the introduction of the ImageCLEF Med-VQA 2018 challenge~\cite{HLF+18}. However, the field is still at a nascent stage and there is still much progress to be made before Med-VQA systems are ready to be deployed in real clinical settings.

The field of medicine has seen great technological advancements with increases in the amount and accessibility of data. With the federal regulations around electronic health records (EHR), patients have now more access to their medical data than ever before. Given that patients can independently check their health records outside of their official consultations, there is an increased need for an accessible way to have their questions answered correctly. Patients can book a consultation with a doctor to obtain answers to their questions, but may be hesitant due to time and money constraints. On the other hand, patients have the option to rely on search engines and conversational agents such as Chat-GPT~\cite{radford2018improving}. However, there is an increased risk of getting misleading or incorrect information. To overcome these challenges, there is a need for a system that helps patients to better manage and understand their medical data without oversight from a healthcare professional. A Med-VQA system could fulfill this need, particularly as its inclusion of natural language question input makes it suited for answering unguided natural language questions, like those that patients may have about their medical images. 

Most existing studies on Med-VQA rely on Convolutional Neural Networks (CNNs) for images and Recurrent Neural Networks (RNNs) for questions and answers as their building blocks. However, less attention has been paid to why a particular component was chosen over another, and why the base model was built in a particular way. There exists a trend in component choice towards more advanced modules without much clarity on why. Hence, it is important to investigate and understand the benefits (or lack) of advanced modules that are being adopted. Datasets for Med-VQA are small, presenting a challenge for a model attempting to learn patterns from them. Therefore, pretraining plays a crucial role in improving model performance. However, medical images and text can be very different in nature from general domain images and text. We hypothesise that pretraining on the medical domain instead could provide performance benefits, since the model is fine-tuned with more domain-specific knowledge that is more directly applicable to the task. Apart from Med-VQA model performance comparison, it is important to have evidence verification, involving visualisation techniques to understand why a particular model gives a particular output for a particular input, and it is crucial to unbox deep learning models in the medical field. Evidence verification is particularly important for a verified Med-VQA system, as it could be making diagnostic judgements about patient's medical images. In this paper, we make the following contributions:
\begin{enumerate}
    \item We systematically compare various components forming the image encoder, question encoder and answer encoder. This helps us to obtain a well-optimised Med-VQA model.

    \item We evaluate the importance of domain-specific knowledge for the Med-VQA task. We not only used pretraining for images but also used pretraining for questions and answers, thereby forcing the Med-VQA model to utilise medical domain knowledge compared to pretrained components in the general domain. 

    \item We use evidence verification techniques to evaluate results. Specifically, we use Gradient Weighted Class Activation Mapping (GradCAM), highlighting regions of the image that are important for predicting a particular answer. 
\end{enumerate}

The rest of the paper is organised as follows: in section 2 related work is briefly reviewed; in section 3 details about the methodology are provided; in section 4 experimental results are presented. In section 5 the results and ablation studies are discussed. Finally, section 6 concludes this paper and recommends future directions.

\section{Related Work}~\label{related_work}

\emph{Deep Learning} (DL), a sub-field of Machine Learning (ML), makes use of neural networks, which are complex models consisting of interconnected units (``neurons") that aim to mimic the human brain in order to learn complex tasks~\cite{Goodfellow:2016:DL_book}. DL gives systems the ability to extract relevant features automatically from the raw data and to be trained in an end-to-end manner. Over the past few decades, researchers have sought to push boundaries within individual fields such as CV and NLP. However, with the rise of DL, researchers found that DL has the advantage of being generalisable to a variety of tasks within a variety of fields. Therefore, researchers started focussing on problems that lie at the intersection of various fields such as CV and NLP, such as \emph{Med-VQA}.

The most common approach for a Med-VQA system is the \emph{joint-embedding framework}. This framework consists of four components: an \emph{image encoder} to extract visual features from an image, a \emph{question encoder} to extract textual features from the question, a \emph{feature fusion algorithm} to combine visual and textual features in a meaningful way and an \emph{answer generation module} to predict an answer in the form of natural language. CNNs are specialised neural networks suitable for processing images or videos, and are typically used as an image encoder. Of the various CNNs, VGG-Net~\cite{SZ14} and ResNet~\cite{HZRS16} have been widely used in Med-VQA systems. Other options include Inception-Net or even an ensemble of different CNNs. For e.g., Gong \emph{et al.}, noticed that the VQA-RAD dataset consists of CT, MRI and X-rays. They trained three separate ResNet models, one on each modality, and then selected the best network for a given input. However, deeper networks showed an overfitting problem due to the increase in model complexity and lack of training data. To compensate for the relatively smaller dataset sizes in Med-VQA, the use of \emph{transfer learning} or \emph{pretrained CNNs} has been widely adopted for Med-VQA systems. However, general domain images are very different to medical images in terms of their features. 

On question answering, Recurrent Neural Networks (RNNs), which are specialised neural networks suitable for processing sequential data such as text and speech, have been widely used. Although vanilla RNNs can successfully remember information with short-term dependencies, they struggle to remember information from further in the past (long-range dependency). To overcome this issue, Long Short-Term Memory (LSTM)~\cite{HS97} and Gated Recurrent Unit (GRU)~\cite{Cho:2014:GRU} networks have been proposed. To further improve the long-range dependency problem, the attention mechanism was introduced and is widely used in conjunction with LSTM or GRU~\cite{Bahdanau:2015:NMT}. The \emph{Transformer}~\cite{Vaswani:2017:Transformer}, an encoder-decoder architecture that is entirely based on the attention mechanism, has recently been the preferred choice for text modelling. Some studies (e.g.~\cite{LWS+20}) chose to discard complex language encoding and make use of light language encoding, such as template matching. This strategy became a popular choice for the VQA-Med 2020 and VQA-Med 2021 challenges~\cite{LZT+21}. Researchers found that questions in these datasets have a repetitive format and belong to only a single category (abnormality), and therefore require only a simple language encoder. For e.g., Liao \emph{et al.} used Skeleton-based Sentence Mapping~\cite{LWS+20}, creating a limited number of templates based on similar questions. However, this method has a clear limitation in that it requires a limited number of question types in order to work well. Bidirectional Encoder Representations for Transformers (BERT)~\cite{DCLT18} based on the Transformer model has also been widely applied for question encoding in  Med-VQA. 

The choice of the fusion algorithm can extend from a simple pooling mechanism to complex attention mechanisms. Common fusion algorithms include  simple concatenation, element-wise multiplication or element-wise sum of image and question features. However, studies showed that simple fusion algorithms are not expressive enough to capture complex associations between image and text, and the outer product of vectors should be used instead. Calculating the outer product is computationally very expensive, therefore methods such as Multimodal Compact Bilinear Pooling have been proposed. Attention mechanisms are much more complex than simple fusion, but can provide better performance as they aim to more meaningfully relate the image and question vectors. One commonly used attention mechanism is the Stacked Attention Network (SAN)~\cite{YHG+16}. SAN has multiple attention layers that interact with the image features multiple times. Each time, the network generates an attention distribution over the image, and adds this to the query vector to generate a “refined” query vector. This allows the network to infer the answer progressively by gradually filtering out irrelevant regions. 

The answer generation component aims to output the correct answer, given the fused image and question features. This can be implemented via classification or generation. In the classification method, models use a softmax layer that outputs one of a finite number of possible answers, whereas the generation method involves using a language decoder to output the correct answer, such as an RNN or Transformer decoder. The classification method is much simpler than the generation method, and works particularly well when the questions and answers are closed-ended, repetitive and therefore limited in number. However, it is clearly more rigid than the generation method, which can become an issue when questions and answers are more complex.

\section{Methodology}
\subsection{Dataset}

\begin{figure}%
    \centering
    \subfloat[\centering Modality category example]{{\includegraphics[height=2cm, width=4cm]{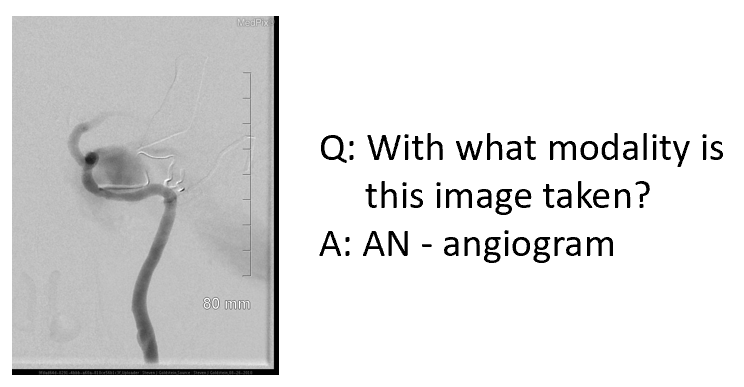} }}%
    \quad
    \subfloat[\centering Plane category example]{{\includegraphics[height=2cm, width=4cm]{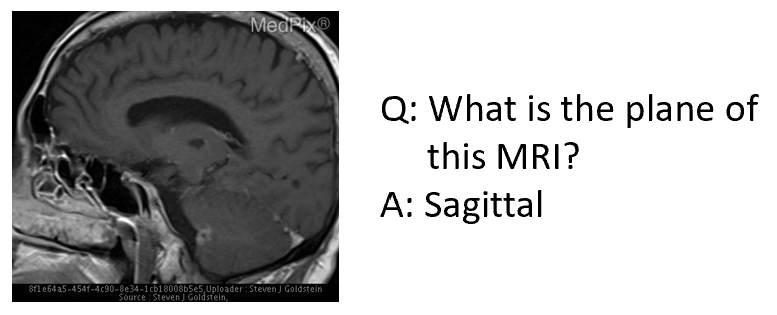} }}%
    \\
    \subfloat[\centering Abnormality category example]{{\includegraphics[height=2cm, width=4cm]{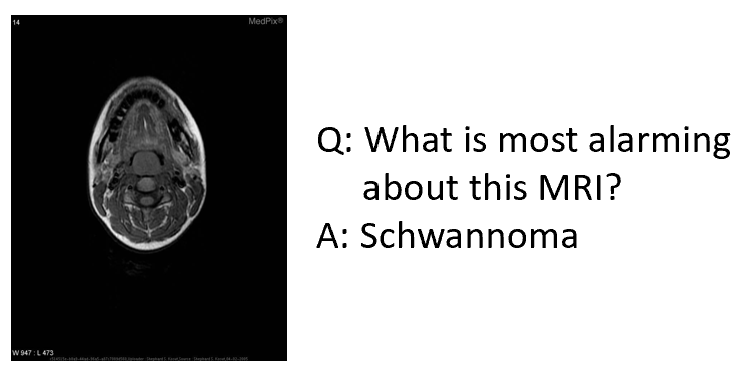} }}%
    \quad
    \subfloat[\centering Organ system category example]{{\includegraphics[height=2cm, width=4cm]{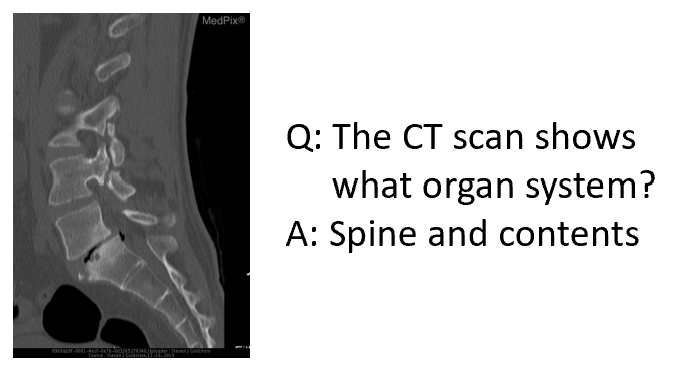} }}%
    \caption{Example data from the VQA-Med 2019 dataset in each of the four categories.}%
    \label{fig:data_sample}%
\end{figure}

We use the VQA-Med 2019 challenge dataset\cite{AHD+19} as it is the largest dataset currently available for the Med-VQA task and has diversity in terms of question categories and image modalities. The dataset consists of 4,200 images from the MedPix database, with 15,992 corresponding question and answer pairs. In Figure~\ref{fig:data_sample} a random set of examples from the dataset is shown. There are four possible question categories---modality, plane, abnormality and organ system.

\begin{figure}
    \centering
    \includegraphics[scale=0.9]{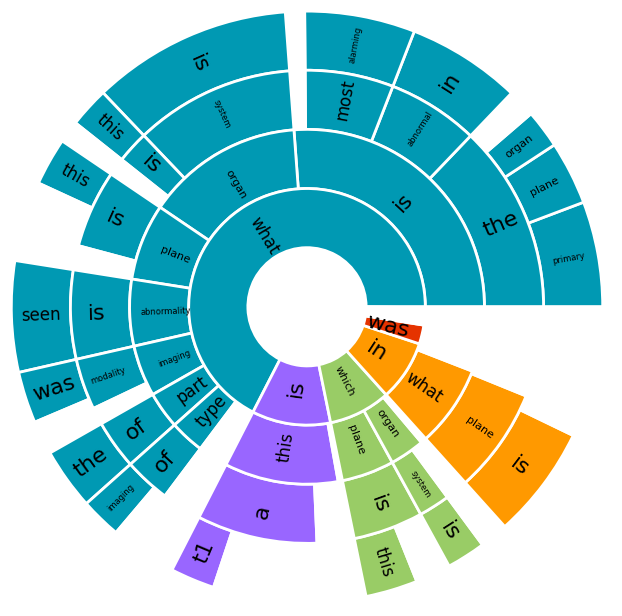}
    \caption{Distribution of the first four words in the VQA-Med 2019 dataset.}
    \label{fig:q_structure}
\end{figure}

The questions in the VQA-Med dataset are generated artificially and then verified by humans. This allows data to be created faster and more cost-effectively, however it also introduces limitations as questions are likely to have less variation in structure. In Figure~\ref{fig:q_structure} a graphical representation of the frequency of each possible word for the first four words of all questions in the dataset is provided. The innermost ring shows the distribution of the first word, the second ring splits this further by the next word and so on. Sections of the chart in white indicate the next words that make up less than 2\% of questions. This chart shows that questions are rigid in structure, with many questions beginning the same way and appearing predominantly close-ended. Most questions begin with ``is this..." or ``which plane...", which implies only one correct answer. This is further evident by examining the number of words in answers. More than 50\% of answers consist of only one word, and more than 82\% of answers have between one and three words. The shortness of answers indicates that there is not much opportunity for them to be worded differently. Based on this analysis, we conclude that the best answer generation strategy is likely to be classification rather than generation, since generation is more suited for open-ended questions.

\subsection{Model development}

\begin{figure*}
    \centering
    \includegraphics[scale=0.7]{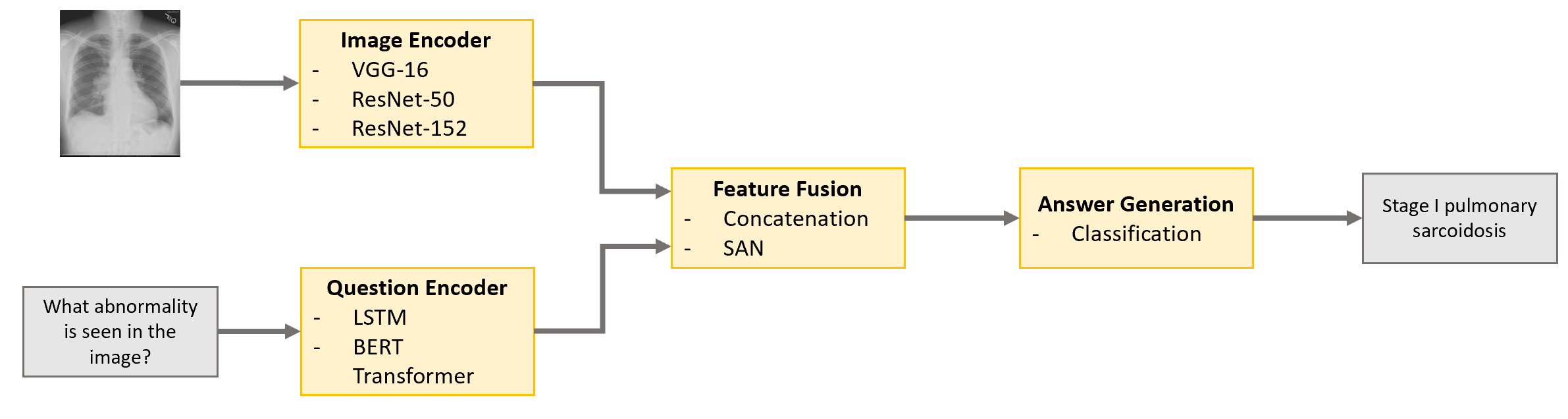}
    \caption{An overview of the structure of our model and the various components tested.}
    \label{fig:model}
\end{figure*}

We use a joint-embedding framework as the structure of our model, and test the performance of various components. An overview of the model structure as well as components tested can be seen in Figure \ref{fig:model}. 

The baseline model was constructed with simpler modules that are commonly used for Med-VQA, even though they may not be the most modern or advanced techniques available. This was done for the preliminary model in order to use methods that are well-tested, to provide a benchmark against which we can measure future experiments. The image is passed through a VGG-16 network that is pretrained on ImageNet to generate a 1-dimensional encoding of the image. As discussed in Section II, VGG-16 is a CNN that is quite straightforward in structure, with thirteen convolutional layers, five max-pooling layers and three dense layers. The question is first tokenised using pretrained word embeddings (specifically, we use BioWordVec embeddings \cite{ZCY+}), which encodes the question as a numerical vector. This is then passed through an LSTM network to generate a 1-dimensional encoding of the questions. These vectors are then concatenated, and this final feature vector is passed through two fully connected layers to generate the final output class.

To improve performance over the baseline, we considered other components that could be used for the image encoder, question encoder and fusion algorithms in the model. These substitutions are summarised in Table \ref{tab:components}.

\begin{table}
\centering
\caption{Component substitutions to be tested}
\begin{tabular}{|c||c|c|}
    \hline
   \bf{Module}  & \bf{Baseline} & \bf{Alternative Component} \\
   \hline \hline
   \bf{Image Encoder}  &  VGG-16 & ResNet-50, ResNet-152 \\
   \hline
   \bf{Question Encoder} & LSTM & BERT Transformer \\
   \hline 
   \bf{Feature Fusion } & Concatenation & Stacked Attention Network \\
   \hline
\end{tabular}
\label{tab:components}
\end{table}

ResNet is a natural substitution to make for the image encoder component, as it is a newer, more advanced network than VGG-Net, that seeks to solve the vanishing gradient problem and allow for much deeper networks. ResNet models achieve a higher accuracy than VGG-Net on the ImageNet classification task \cite{HZRS16}. 

For the question encoder, we tested the performance of a BERT transformer compared to the LSTM network that was used for the baseline model. The transformer discards the complex recurrent structure that was used by the LSTM and other similar NLP models, instead using only attention mechanisms to process text. BERT is pretrained on English Wikipedia and BookCorpus for the language modelling task. 

In the baseline model, the image and question feature vectors were fused by concatenating the two vectors. This is one of the simplest methods, relying on the fully connected layers in the final answer generation component to form meaningful connections between the question and image feature vectors. However, as discussed before, using an attention mechanism could be a better way to fuse image and question feature vectors. We applied Stacked Attention Network (SAN), which uses multiple attention layers to progressively refine attention distributions over the image, in order to focus on parts of the image that are more relevant to the question. This could provide benefits to the model's understanding of the image as it relates to the given question. To implement SAN, we discarded the fully connected layers from the image encoder to maintain positional information in the encoding, and pass computed image and question features to the SAN, rather than just concatenating them. We tested SAN with 2, 3, or 4 attention layers, achieving best results using 3 attention layers. 

\subsection{Incorporating medical domain knowledge}

One issue that has been integral to applying deep learning models on the Med-VQA task is the lack of large-scale annotated datasets. Therefore, it is important to consider techniques that could help mitigate this issue. Both the image encoder (VGG-Net or ResNet) and language encoder (BERT) can be pretrained on general domain images or text, respectively. However, although this pretraining is invaluable, medical images and medical text in Med-VQA are undoubtedly different from the general domain. Incorporating medical domain knowledge as part of pretraining could help the model to learn representations that are more directly applicable to downstream tasks, leading to improved performance. We implemented pretraining for both the question encoder and the image encoder to evaluate the benefits of using medical-specific pretraining for the Med-VQA task. 

For the image encoder, given that there is no large-scale annotated dataset available, we used \emph{self-supervised pretraining} which involves training the network on unlabelled data through tasks that allow it to learn a generalised representation of images. Of the various methods available for self-supervised pretraining, we applied the \emph{contrastive learning} method, similar to the method implemented by SimCLR \cite{CKNH20}, although modified for application to the VGG-Net encoder and using data augmentations that are more applicable to the medical image dataset. We used the Radiology Objects in COntext (ROCO) dataset~\cite{PKR+18}, which consists of over 81,000 radiology images in a wide variety of imaging modalities. ROCO was chosen as it is large, diverse and has images similar to the Med-VQA task. The image encoder was pretrained for 80 epochs, with a batch size of 128. 

The original BERT transformer uses general domain BERT pretraining, which pretrains the model on two NLP tasks (language modelling and next-sentence prediction). The pretraining corpus consists of BooksCorpus (an approximately 800 million word collection of freely available novels) and English Wikipedia (approximately 2.5 billion words). However, since medical language can be very different from general domain language, using a significant amount of specialised terminology, it is thought that giving the transformer a better understanding of medical language could improve its performance. To investigate this, we used a BERT model called \emph{BioBERT}~\cite{LYK+20}, pretrained on the same tasks as BERT, but using the PubMed corpus, which comprises more than 35 million citations for biomedical literature from MEDLINE, life science journals, and online books (approximately 4.5 billion words). This makes BioBERT particularly suited to biomedical NLP applications. 

\subsection{Evidence Verification}

In our experiments, we used classification accuracy to evaluate the performance of the model. However, quantitative evaluation cannot evaluate the quality of model reasoning. Evidence verification involves generating output that gives insight into why the model generated a particular answer, and it is crucial for deep learning models in the medical field, particularly when they may be making diagnostic judgements. We used Gradient Weighted Class Activation Mapping (GradCAM) \cite{SCD+17} for evidence verification. GradCAM uses the gradients of a target output class flowing into the final convolutional layer of a network to produce a localisation map that highlights regions of the image that were important to predicting that class. In this way, GradCAM can produce a heat map over the input image showing the areas the model paid the most attention to in order to produce its answer. To implement GradCAM, we used a Python package\footnote{\url{https://github.com/jacobgil/pytorch-grad-cam}}, with some modification in order to handle the two multi-modal inputs that are required for our network. We then qualitatively examined the heat map outputs.

\section{Results}~\label{results}

All experiments were implemented in Python using the PyTorch library\cite{PGM+19}. To ensure robustness of results, we performed five-fold cross-validation on the dataset. This was done by randomly splitting the dataset into training set (80\%) and test set (20\%), and repeating the process five times. The split was seeded with the same number for all versions of the model to enable fair comparison. The model was trained for 50 epochs, optimising loss with the Adam optimiser \cite{KB14} with a learning rate of $1e-4$ and a batch size of 64. The input images were normalised and data augmentation was performed during training to increase dataset size and minimise model overfitting. We changed image brightness and contrast by 5\% with a probability of 0.4, translation and rotation by 5 units with probability of 0.5, adding Gaussian blur with probability of 0.5 and adding Gaussian noise with a probability of 0.4. 

\subsection{Quantitative results}

The baseline model achieved an accuracy of $0.56\pm0.01$. We can see that there is still some overfitting happening, despite the data augmentation. Qualitatively examining the model's outputs, there is a marked difference in the model's ability for different question categories. The model has reasonable accuracy on plane and organ system questions (78\% and 74\% respectively), and a lower accuracy on modality questions (64\%). However the model's performance on abnormality questions is extremely poor, at only 6\%. This is as expected, since in this dataset most abnormality classes have very few examples, making it very difficult for the model to learn to recognise them. A limitation of the accuracy metric is that it cannot account for cases where the model was technically correct but gave the answer in different wording. For example, in a question asking ``What is abnormal in the CT scan?" the model answers ``Pulmonary embolism", where the ground truth answer is ``PE", an acronym that stands for pulmonary embolism. These types of issues can only be fixed by someone with professional medical knowledge updating the dataset to make these answers consistent. However, even accounting for these cases, the model's performance on abnormality questions is still extremely low compared to other question categories. 

\begin{table}
    \centering
    \caption{Test accuracy achieved by each model variation.}
    \begin{tabular}{lc}
        \toprule
          \bfseries Model Variation  & \bfseries Test Accuracy \\
        \midrule
          VGG-16 + LSTM + Concatenation  & 0.56 \\
          ResNet-50 + LSTM + Concatenation  & 0.54 \\
          ResNet-152 + LSTM + Concatenation  & 0.53 \\
          \bfseries VGG-16 + BERT + Concatenation  & \bfseries 0.60 \\
          VGG-16 + BERT + SAN  & 0.58 \\
          \bfseries VGG-16 + BioBERT + Concatenation & \bfseries 0.60 \\
          \bfseries Pretrained VGG-16 + BERT + Concatenation & \bfseries 0.60 \\
    \bottomrule
    \end{tabular}
    \label{tab:results}
\end{table}

In Table~\ref{tab:results} the overall test accuracy achieved by each of the model variations as detailed in Section III are shown. Overall best performance is achieved using a VGG-16 image encoder, BERT Transformer question encoder and concatenation as the feature fusion strategy, achieving a test accuracy of $0.60\pm0.01$. For the image encoder, we find that deeper and more complex networks such as ResNet-50 or ResNet-152 do not provide better results and in fact demonstrate a higher degree of overfitting. This clearly highlights the issue of the small dataset size of Med-VQA. For question encoding, the BERT transformer gave higher test performance compared to the LSTM network. 

In Table~\ref{tab:results_breakdown} the test accuracy is shown by question category for the baseline model compared to the model with highest test accuracy, showing in more detail exactly what aspects the latter model improves on. There is a significant increase in accuracy in the modality category (+12\%) and a modest increase in the abnormality category (+2\%), which allows the +BERT model to achieve an increased overall accuracy. 

\begin{table}[]
    \centering
    \caption{Accuracy of the baseline versus BERT model per category type.}
    \begin{tabular}{lcc}
    \toprule
        \textbf{Model Variation} & Baseline & +BERT  \\
    \midrule
    Modality & 0.64 & 0.76 \\
    Plane & 0.78 & 0.77 \\
    Organ & 0.74 & 0.74 \\
    Abnormality & 0.06 & 0.08 \\ 
    \hline
    Overall & 0.56 & \textbf{0.60} \\
    \bottomrule
    \end{tabular}
    \label{tab:results_breakdown}
\end{table}

\begin{figure*}%
    \centering
    \subfloat[\centering Category misclassification]{{\includegraphics[height=3.5cm, width=3.5cm]{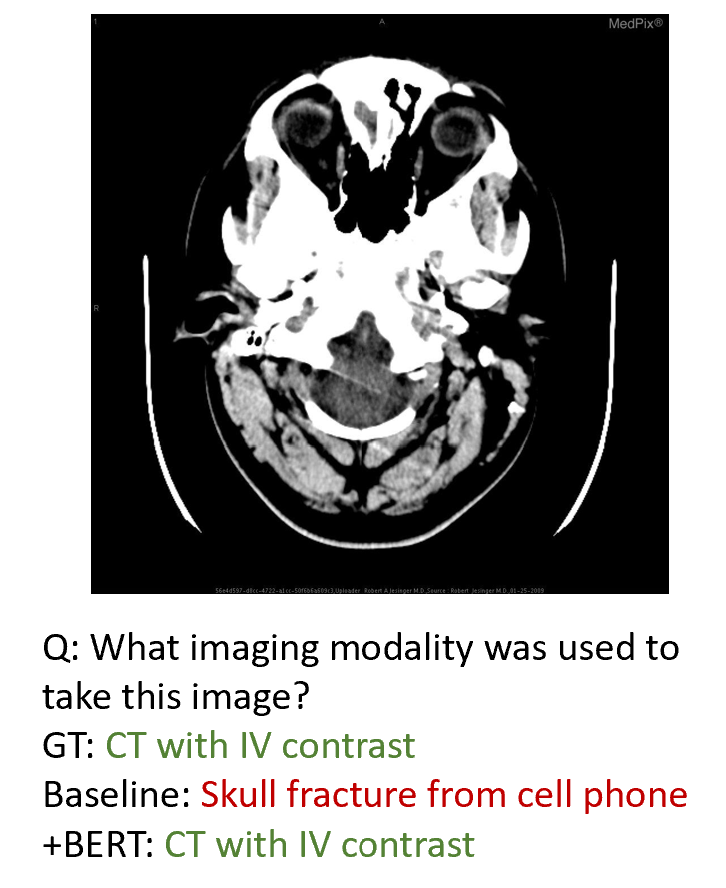} }}%
    \quad
    \subfloat[\centering Wrong answer type]{{\includegraphics[height=3.5cm, width=3.5cm]{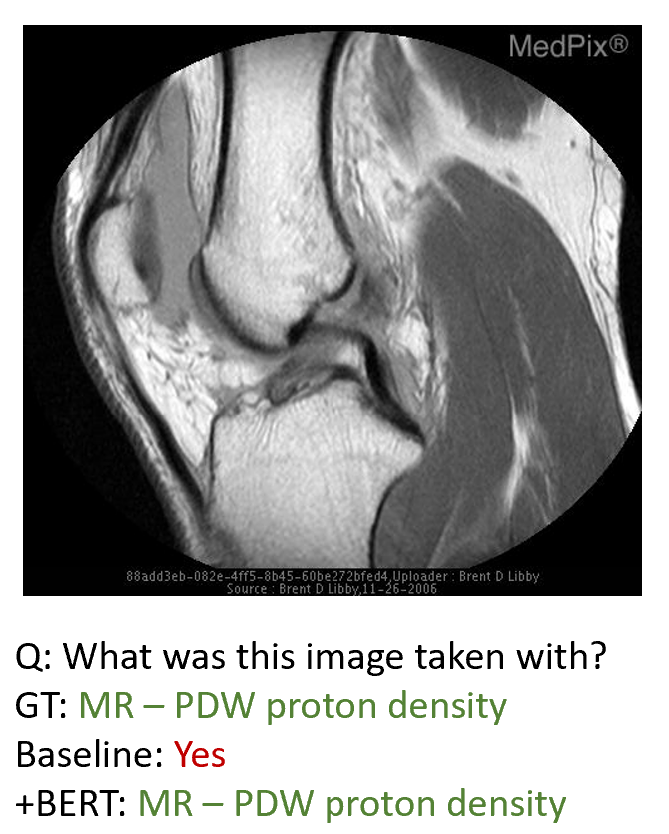} }}%
    \quad
    \subfloat[\centering Not choosing from options]{{\includegraphics[height=3.5cm, width=3.5cm]{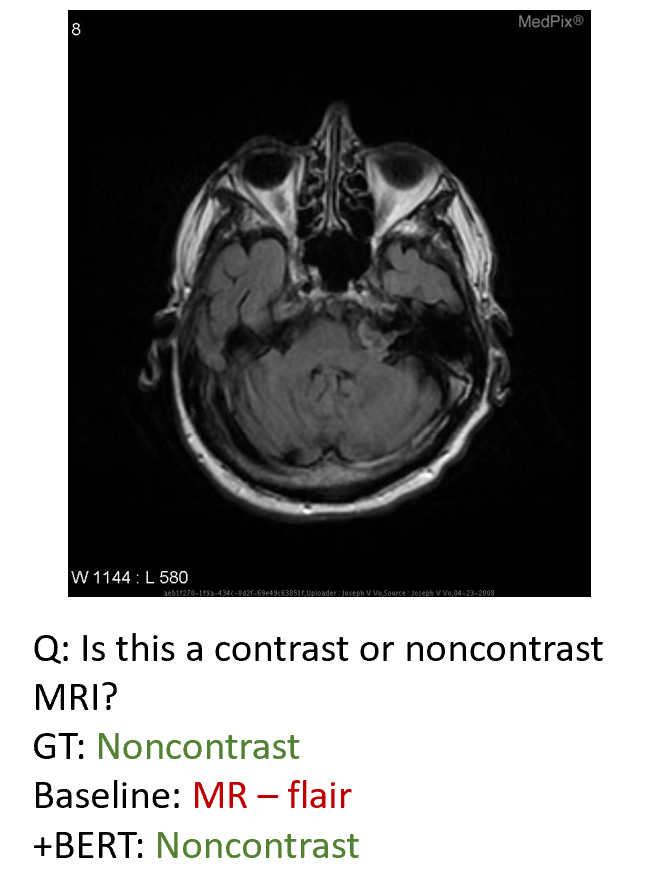} }}%
    \caption{Example responses from LSTM and BERT variations of model.}%
    \label{fig:bert_ex}%
\end{figure*}

By examining the model's answers, we find that this improvement is due to the model's improved understanding of the question. For example, the model no longer misclassifies question categories. In the baseline model, approximately 8\% of abnormality questions were misclassified as other categories by the model, whereas the BERT model now correctly identifies all abnormality questions (e.g. Figure ~\ref{fig:bert_ex}a), providing a performance benefit. Secondly, the BERT model is better able to understand the required answer type, and always answers questions in a way that makes sense. For example, in Figure~\ref{fig:bert_ex}b, the baseline model incorrectly identifies this question as requiring a yes/no answer, whereas the BERT model is able to give a reasonable answer to the question. Similarly, as shown in Figure~\ref{fig:bert_ex}c, the baseline model would sometimes incorrectly handle questions requiring the model to select from the given options, whereas the BERT model always chooses from the provided options for these questions. These improvements show that a better understanding of the question can lead to higher accuracy overall. 

\begin{figure*}%
    \centering
    \subfloat[\centering]{{\includegraphics[height=3.5cm, width=4.5cm]{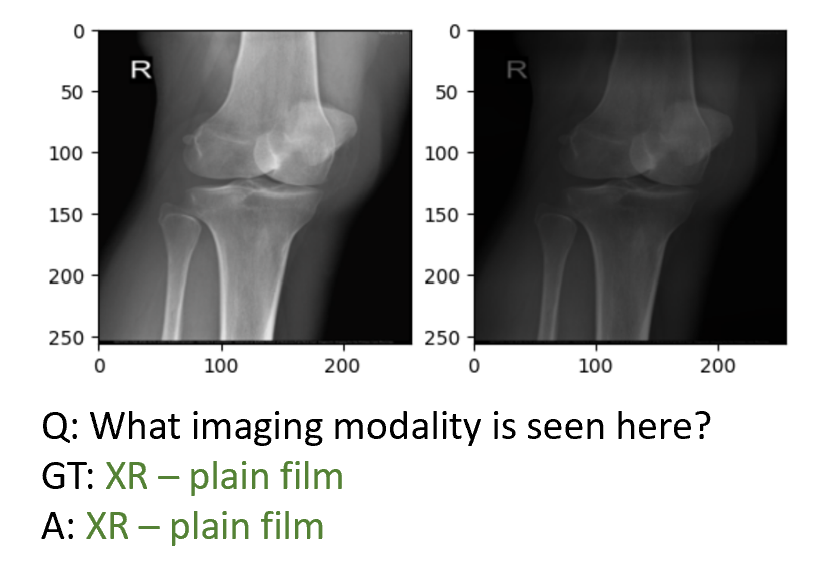} }}%
    \quad
    \subfloat[\centering]{{\includegraphics[height=3.5cm, width=4.5cm]{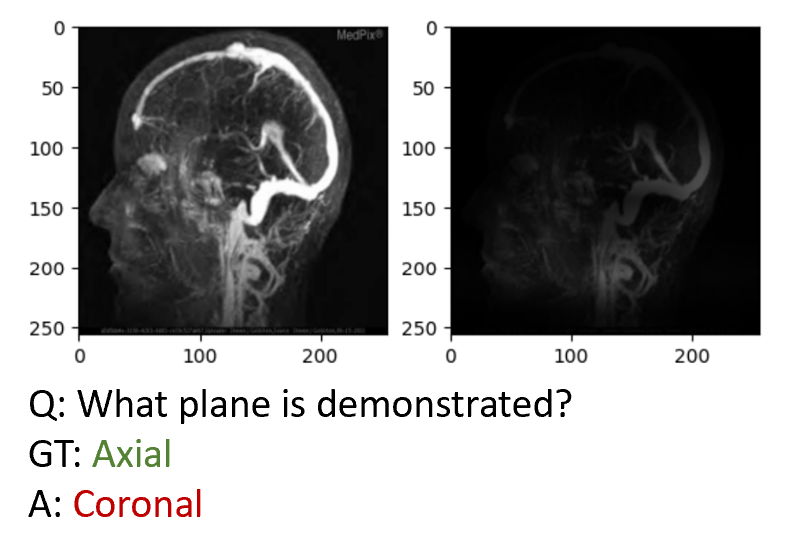} }}%
    \quad
    \subfloat[\centering]{{\includegraphics[height=3.5cm, width=4.5cm]{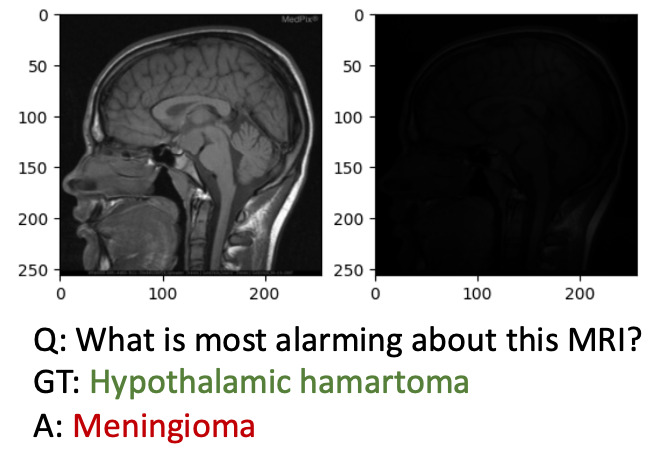} }}%
    \caption{Examples of attention distribution output by SAN fusion method.}%
    \label{fig:attn_ex}%
\end{figure*}

Our results in Table~\ref{tab:results} show that SAN as a feature fusion method did not improve results compared to the concatenation method. This is because there is simply not enough data for the model to learn a useful refined attention distribution, and the added complexity of generating an attention distribution can actually cause the model to obscure some useful information. By qualitatively examining the model's generated attention distributions (examples of which are shown in Figure~\ref{fig:attn_ex}, with original images on the left, and attention distributions on the right), we find that the issue is that the questions in the dataset need either a holistic view of the image (for example most questions in the modality, plane and organ system categories), or a very strong focus on a small part of the image (for example most abnormality questions). For the former type, we find that the model either attends to all parts of the image more or less equally (e.g. Figure~\ref{fig:attn_ex}a), making attention redundant, or the attention distribution obscures relevant parts of the image (e.g. Figure~\ref{fig:attn_ex}b). For questions requiring good localisation, the model instead tends to produce attention distributions of the type shown in Figure~\ref{fig:attn_ex}c, where the model does not attend to any part of the image in particular. This is due to insufficient examples of particular abnormalities in the dataset, therefore the model is not able to learn the small areas of interest in these images, and instead does not focus on any part, leading to performance loss. 

The results in Table~\ref{tab:results} show that neither domain-specific pretraining method that we tested provided a performance benefit. Our testing of BioBERT as the question encoder showed that medical domain-specific pretraining does not appear to be important to the question encoder for the VQA-Med 2019 dataset. This is likely because of the lack of medical language used in the questions of this dataset. As previously noted, the questions in this dataset are rigid in structure, likely generated from a limited number of templates. Almost all medical language in the text comes from the answers, rather than the questions. Although there are some exceptions (e.g. ``angiogram", ``gastrointestinal", ``ultrasound"), these medical terms are never crucial to understanding the question as a whole. Therefore, using a question encoder that has been pretrained on the medical domain does not appear to help the model better understand the questions in this dataset. 

Self-supervised pretraining for the image encoder also did not appear to improve the final performance of the model. This is likely due to the distinct types of visual reasoning that are required for different questions. Some questions, such as organ system identification, require the model to be able to distinguish between large-scale differences in images. However, other questions, such as abnormality questions, require the model to be able to recognise very small-scale differences between images. Our results indicate that contrastive learning, at least in the generalised form that was tested here, is not beneficial for improving performance on the Med-VQA task, likely because it is not able to aid the model in performing both types of image differentiation.

\subsection{Qualitative results}

\begin{figure*}%
    \centering
    \subfloat[\centering]{{\includegraphics[height=3.5cm, width=3cm]{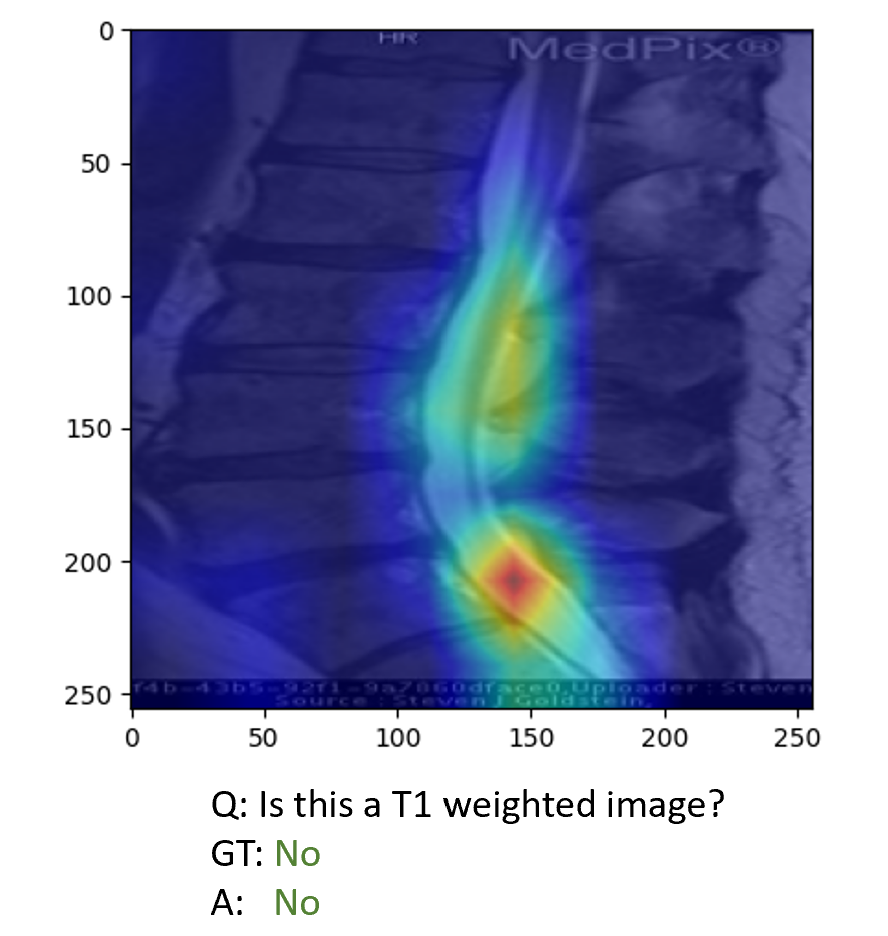} }}%
    \quad
    \subfloat[\centering]{{\includegraphics[height=3.5cm, width=3cm]{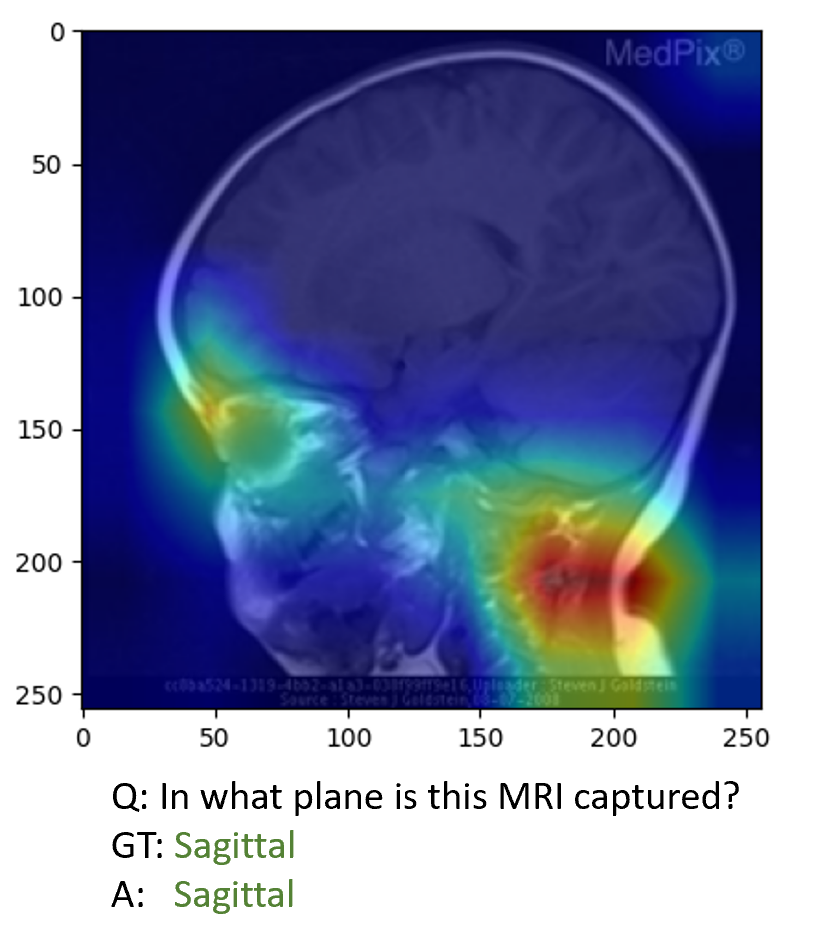} }}%
    \quad
    \subfloat[\centering]{{\includegraphics[height=3.5cm, width=3cm]{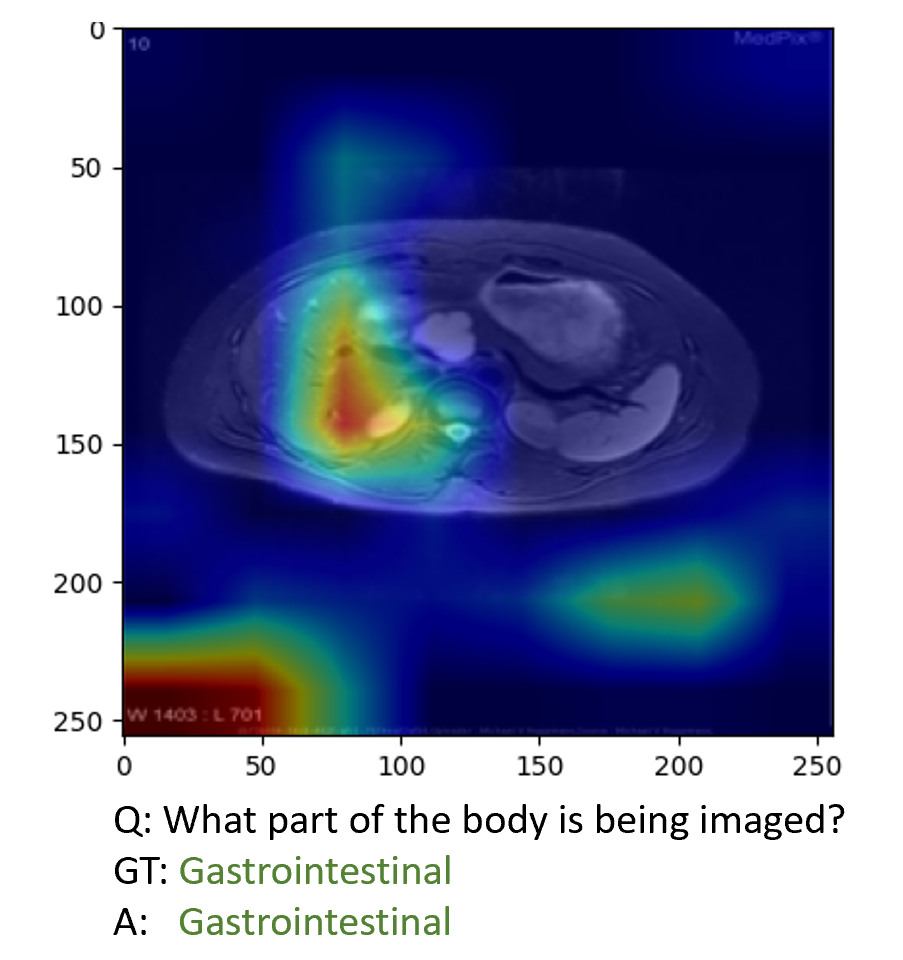} }}%
    \quad
    \subfloat[\centering]{{\includegraphics[height=3.5cm, width=3cm]{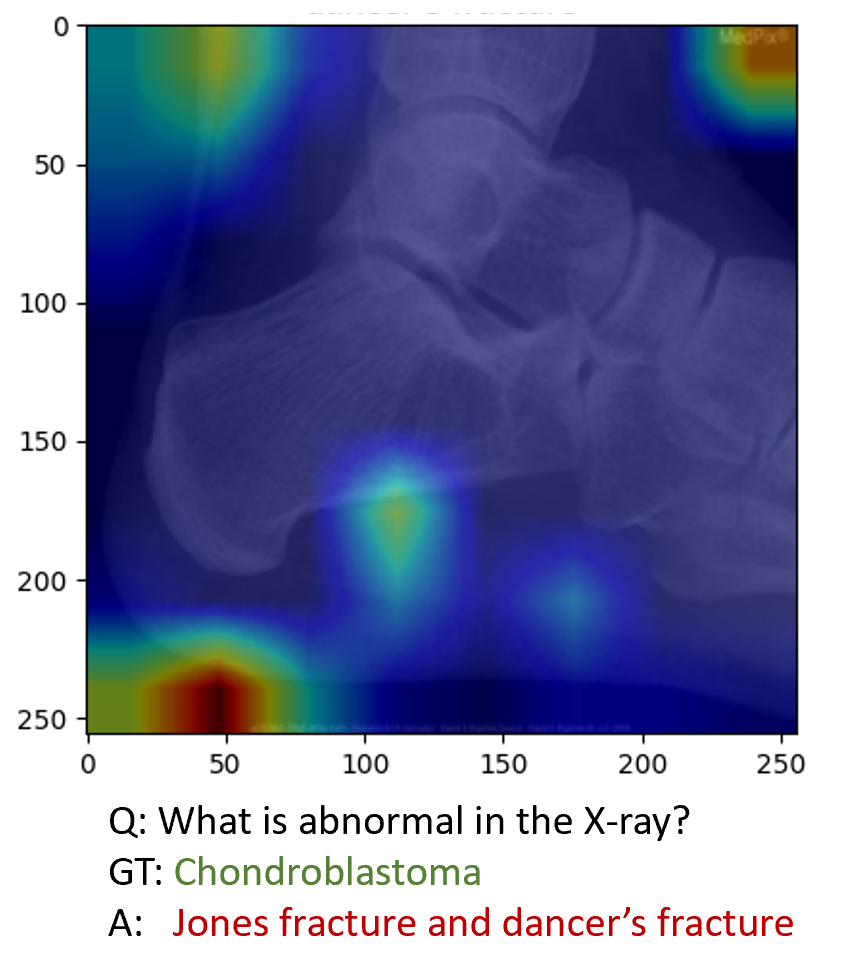} }}%
     \quad
    \subfloat[\centering]{{\includegraphics[height=3.5cm, width=3cm]{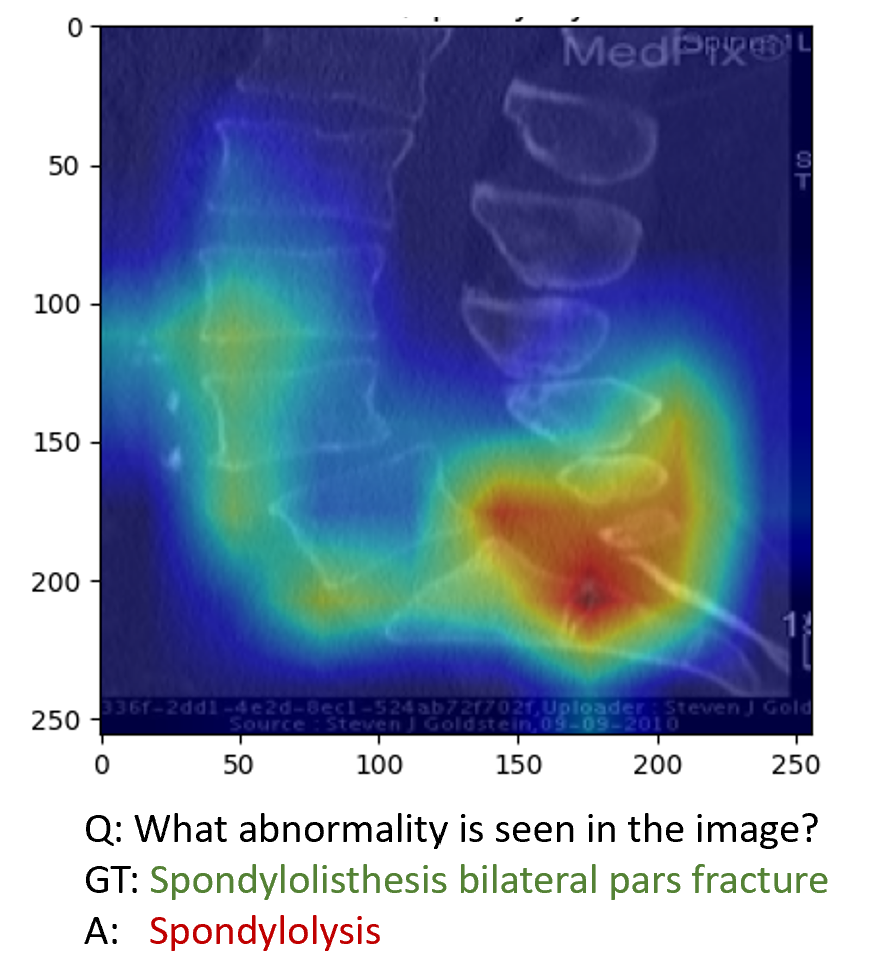} }}%
    \caption{Example output from GradCAM.}%
    \label{fig:gradcam}%
\end{figure*}

In order to evaluate the quality of the model's reasoning when producing answers, we used GradCAM to produce visualisations of the model's attention over the input image, and Figure~\ref{fig:gradcam} shows examples of these outputs.

We found that for the most part, the model does well at predicting the answer from reasonable parts of the image. In Figure~\ref{fig:gradcam}a, we see that when asked to determine whether the image is T1 weighted, the model focusses on the light band in the centre of the image. Since this band is dark in T1 images, this is a good reason for the model to make its prediction. In Figure~\ref{fig:gradcam}b, the model is asked to determine the plane of the MRI, and focusses in particular on the back of the neck and the eye area to produce the correct ``sagittal" answer. These two features would only occur in skull images taken from the sagittal plane, so this is again good reasoning. 

There are however some cases where the model's visual reasoning could be improved. This can either mean that it gives the correct answer for the wrong reason, or it gives the wrong answer because it focusses on irrelevant areas of the image. GradCAM allows us to identify both of these cases, examples of which are shown in Figure~\ref{fig:gradcam}c and Figure~\ref{fig:gradcam}d. In Figure~\ref{fig:gradcam}c, the model correctly identifies the image to be a gastrointestinal image, however the GradCAM output shows that this decision was based predominantly on the presence of the text in the bottom left corner of the image. This indicates that the model has identified this as a shortcut in the data, and has not really learned the correct features. In Figure~\ref{fig:gradcam}d the model provides the wrong answer due to wrong reasoning. The model does give an abnormality as the answer, showing that it has identified the correct question category, but has not selected the correct abnormality. The heat map reveals this to be likely a guess, based on the fact that the model was not focussing on the bone at all. 

\begin{figure}
    \centering
    \includegraphics[scale=0.4]{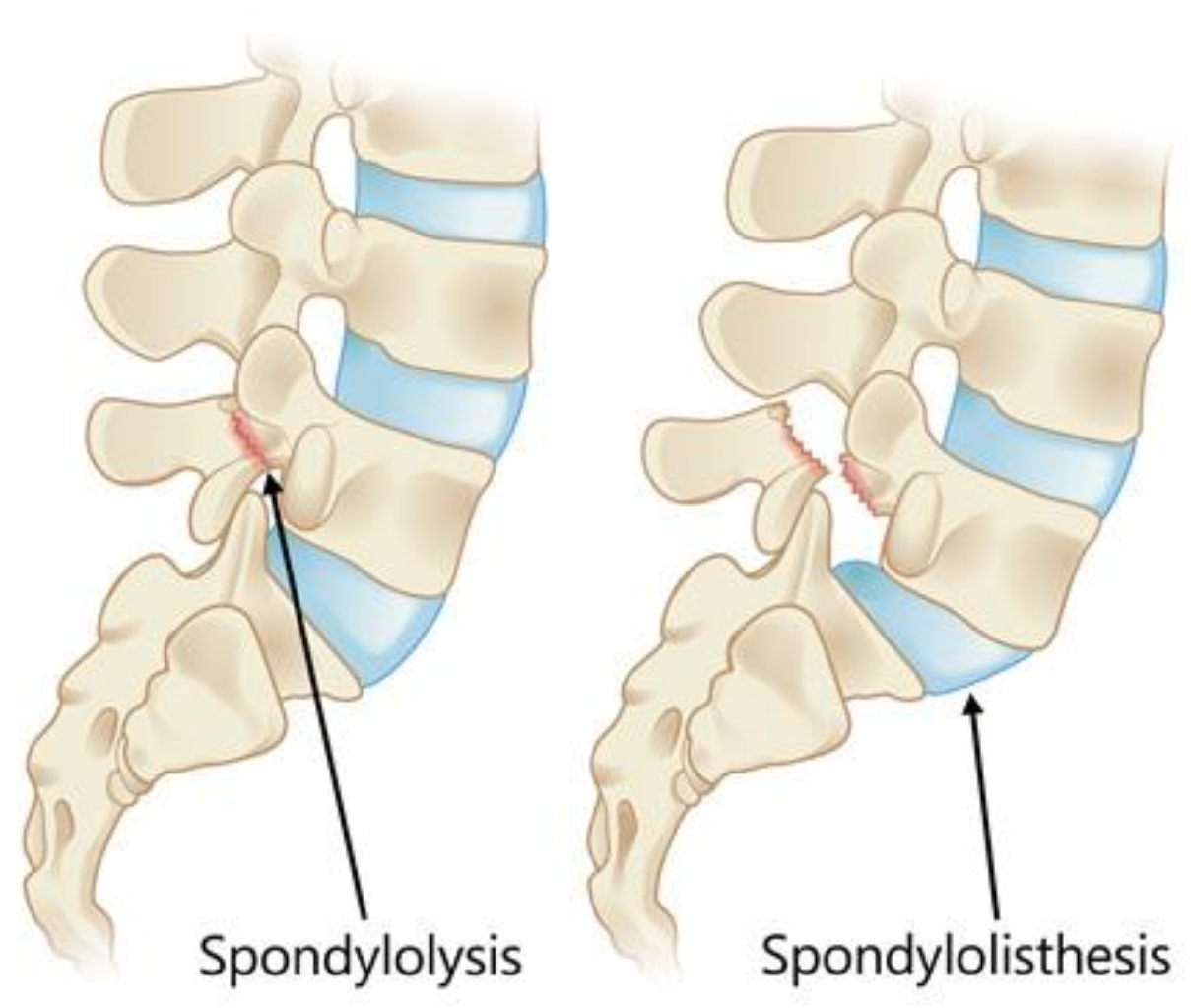}
    \caption{Spondylolysis vs. spondylolisthesis\cite{AAOS20}.}
    \label{fig:spondylolysis}
\end{figure}

Finally, analysis of GradCAM outputs can also identify answers that were almost correct. For example, in Figure~\ref{fig:gradcam}e, the model is asked for the abnormality, and answers with a disorder that is similar but not the same as the ground truth. Comparing the heat map with the diagram in Figure~\ref{fig:spondylolysis} shows that the model was focussing on the correct parts of the image to be able to distinguish between the two disorders. Its failure to predict the correct answer indicates that there was not enough training data for the model to accurately distinguish between the two disorders, but also shows that the model was doing the correct reasoning on the image. With more data, the model would certainly be able to improve its accuracy on abnormality images.

\section{Discussion}~\label{discussion}

In this work, we quantitatively compared the performance of various components that are commonly used for the Med-VQA task, allowing us to evaluate their appropriateness for this task against each other. Our results show that, in general, less complex models with fewer parameters tend to perform better on this dataset. We find that models that achieve better performance in the general domain do not necessarily achieve better performance on the Med-VQA task. The Med-VQA field deals with a low-data regime, and therefore benefits from simpler models that are less likely to overfit.

We conducted domain-specific pretraining in both the question and image encoders. For the question encoder, we found that domain knowledge is not useful for the VQA-Med 2019 dataset, as the questions use limited medical terminology. However, it is possible that a more complex dataset could benefit from question encoder pretraining. For the image encoder, we developed and tested a contrastive learning pretraining method. Our results indicate that this pretraining did not benefit our model, and we propose that Med-VQA is not suited to such a generalised contrastive learning method, owing to the different kinds of visual reasoning required for different questions--some requiring an understanding of large-scale differences, and others requiring an understanding of very small-scale differences. 

We used GradCAM to output heat maps corresponding to the visual attention of our model, allowing us to better evaluate our results by gaining insights into the model's visual reasoning. We find that this provides great benefits to our model's interpretability, allowing us to gain a deeper understanding of the strengths and shortcomings of the model, beyond the quantitative accuracy metric.

One of the biggest challenges inherent to the Med-VQA field is the lack of data. Datasets are very small, but consist of a very wide variety of different possible answers, presenting a barrier to achieving very high accuracy on the task. In this work we investigated self-supervised contrastive learning as a domain-specific pretraining method to mitigate this issue, but found that it did not aid model performance. However, other domain-specific pretraining methods could still be investigated that may help this problem. Further, pretraining alone will not be able to fully overcome this issue. In future, focus should be on generating significantly larger datasets. Future work could involve investigation into automated dataset generation methods for this task, in order to generate bigger and more diverse datasets for training. Better evidence verification techniques and evaluation are also crucial to further development in this area. For medical diagnosis, it is essential to be able to verify why the system answered the way it did, to ensure that the answer is supported by the right evidence. Although we have found GradCAM to give good insight into the visual reasoning of the model, future work could, for example, explore the model's attention on the question features to better evaluate how the model interprets the question. 

\section{Conclusions}~\label{conclusion}

In this work, we conducted an investigation of some existing and new techniques used in the Med-VQA task. We systematically evaluated some common Med-VQA components, and found that generally, simpler and shallower models benefit this task due to the small dataset size. We investigated the importance of domain knowledge in the question encoder and found that textual domain knowledge is not beneficial for this dataset. We developed and tested a contrastive learning pretraining method in the image encoder, and found that the contrastive learning method is not suited to the varied types of visual reasoning required for this task. Our final model achieved a 60\% accuracy on the VQA-Med 2019 dataset. This matches current state of the art models in non-ensembled versions (e.g. \cite{KBM+21}, \cite{RZ20}), suggesting that ensembling our model could provide further performance benefit. We also evaluated our model's results and reasoning qualitatively using GradCAM, finding that as a whole our model is able to show good judgement in making decisions. 

\section*{Acknowledgement}
The authors would like to thank organisers of the ImageCLEF VQA-Med challenge for their time and effort in curation of the MED-VQA dataset, and sharing it for research purposes.  

\section*{Declaration of competing interest}
The authors declare no potential conflicts of interest.

\bibliography{references.bib}
\bibliographystyle{IEEEtran}

\end{document}